\documentclass[a4paper,10pt,conference]{ieeeconf}
\usepackage{cite}
\usepackage{amsmath,amssymb,amsfonts}
\usepackage{graphicx}
\usepackage[utf8]{inputenc}
\usepackage{textcomp}
\usepackage{xcolor}
\usepackage{subfiles} 
\usepackage{amsmath}
\usepackage{url}
\usepackage{algorithm}
\usepackage{multirow}
\usepackage{algpseudocode}
 \usepackage{booktabs}
\makeatletter
\def\endfigure{\end@float}
\def\endtable{\end@float}
\makeatother
\usepackage[caption=false,font=footnotesize]{subfig}
\usepackage{setspace} 
\usepackage{censor}

\usepackage{amsfonts}
\usepackage{hyperref} 
\usepackage{tikz}

\usetikzlibrary{spy, decorations.pathreplacing}

\IEEEoverridecommandlockouts                              

\title{\LARGE \bf
WiFi-based Global Localization in Large-Scale Environments Leveraging Structural Priors from osmAG
}

\author{Xu Ma$^{1}$, Jiajie Zhang$^{1}$, Fujing Xie$^{1}$ and S\"oren Schwertfeger$^{1}$}

\hypersetup{pdftitle={WiFi-based Global Localization in Large-Scale Environments Leveraging Structural Priors from osmAG},pdfauthor={Xu Ma, Jiajie Zhang, Fujing Xie, Sören Schwertfeger},unicode=true,pdfsubject={IEEE ROBIO 2026},pdfkeywords={WiFi localization, osmAG, structural priors}}

\begin{document}

\maketitle
\thispagestyle{empty}
\pagestyle{empty}
  \thanks{}%
\footnotetext[1]{The authors are with the Key Laboratory of Intelligent Perception and Human-Machine Collaboration - ShanghaiTech University, Ministry of Education, China. \{maxu2023, zhangjj2023, xiefj, soerensch\}@shanghaitech.edu.cn \\ This work was supported by the National Natural Science Foundation of China under Grant W2531052 and the Key Laboratory of Intelligent Perception and Human-Machine Collaboration (ShanghaiTech University), Ministry of Education, China. The experiments were supported by the Core Facility Platform of Computer Science and Communication, SIST, ShanghaiTech University.}


\begin{abstract}
WiFi localization offers a lightweight global prior for robots in GPS-denied environments, but fingerprinting requires dense surveys and generalizes poorly beyond surveyed areas. We model received signal strength indicator (RSSI) measurements using OpenStreetMap Area Graph (osmAG) structural priors. Offline, known anchor access points (APs) support sequential calibration of line-of-sight (LOS) propagation and wall attenuation; unknown AP locations are then estimated by alternating ray-wall intersection checks, signal compensation, and least-squares fitting. The resulting AP map supports online robot position estimation using the same structure-aware model. In experiments in an 11,025~m$^2$ multi-floor environment, our method reduces the mean AP localization error from 5.86~m to 3.79~m. Mean robot localization errors are 3.12~m in fingerprinted areas and 3.83~m in unfingerprinted areas, where four baselines yield 11.00--20.21~m. The WiFi-related map representation achieves approximately 251:1 compression relative to the raw fingerprint database. These results support structural modeling for compact global localization beyond surveyed regions. Code and data: \url{https://github.com/XuMa369/osmag-wifi-localization}.

\end{abstract}
\section{INTRODUCTION}
Global localization enables recovery from localization failures, including the \textit{kidnapped robot problem}, in GPS-denied environments. Visual and LiDAR methods provide accurate estimates but can be costly and ambiguous in texture-less, dynamic, or repetitive environments. WiFi offers a lightweight alternative using received signal strength indicator (RSSI) measurements from existing infrastructure~\cite{biswas2010wifi}.

\begin{figure}[t!]
    \centering
    \includegraphics[width=\linewidth]{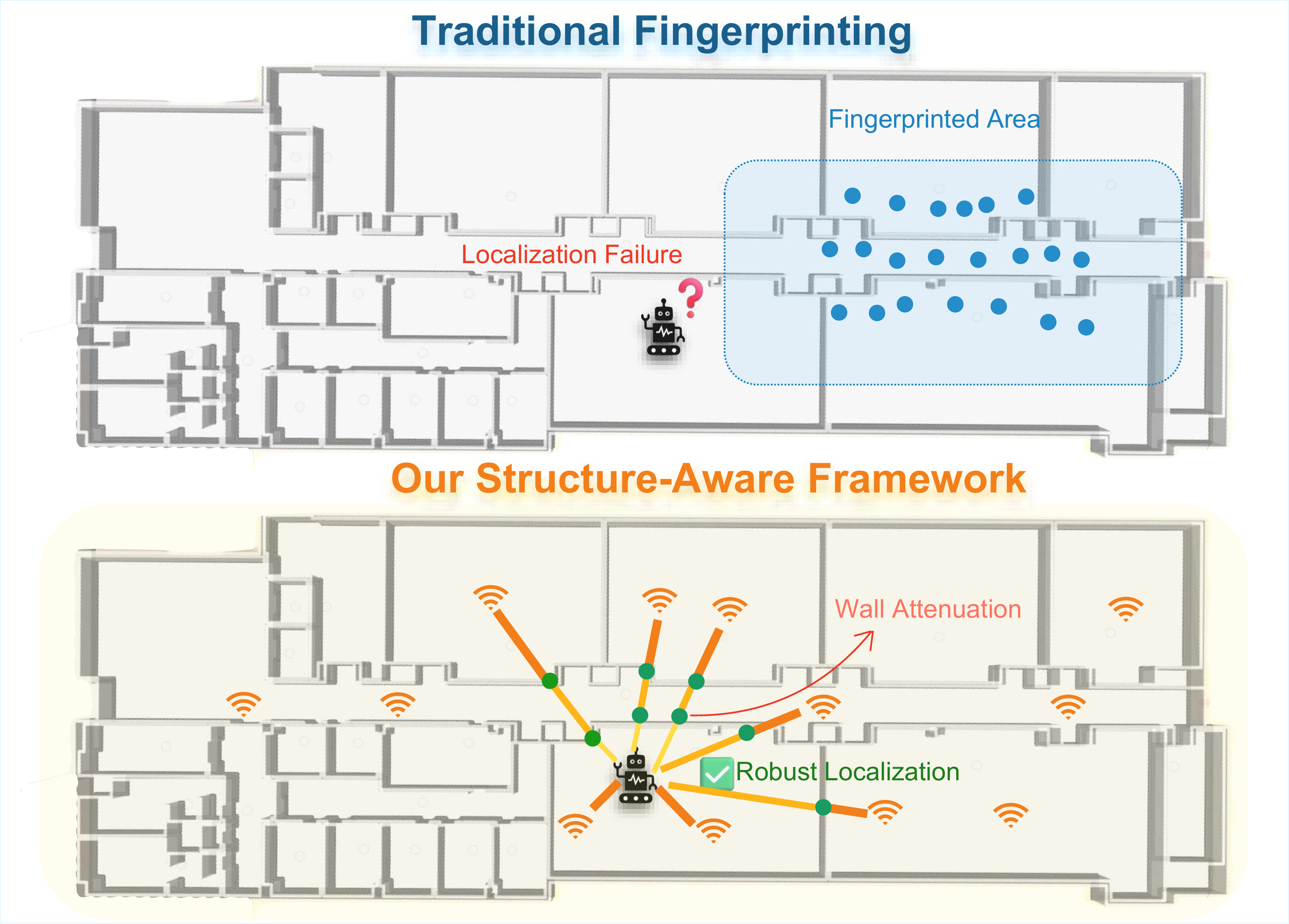}
 
    \caption{\textbf{Beyond surveyed areas.} Top: fingerprinting depends on historical scans and can fail outside their coverage. Bottom: osmAG wall intersections (green nodes) determine attenuation along signal rays, enabling structure-aware localization.}
 
    \label{fig:teaser}
\end{figure}

Fingerprinting requires dense surveys and extrapolates poorly to unvisited areas~\cite{bahl2000radar,10416396}. Gaussian-process methods interpolate sparse observations but smooth over wall-induced discontinuities~\cite{Miyagusuku2016,miyagusuku2018data}; conventional propagation models often omit environmental geometry~\cite{seidel1992914}. Fig.~\ref{fig:teaser} contrasts these limitations with explicit wall modeling.

A useful global prior should remain available when the robot leaves the surveyed footprint. Our approach treats walls as propagation constraints rather than requiring a nearby fingerprint match. It retains the low sensing cost of WiFi while using structural information already available in a semantic indoor map.

\begin{figure*}[t]
    \centering
    \includegraphics[width=0.95\linewidth]{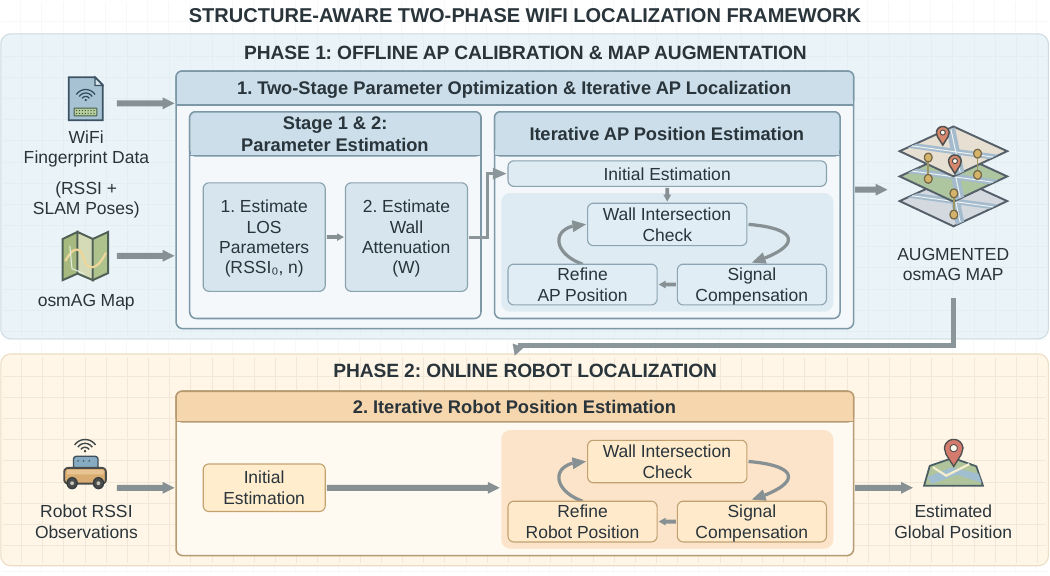}
    \caption{Structure-aware WiFi localization. Offline calibration learns line-of-sight (LOS) and wall-loss parameters from anchor APs, then maps unknown APs. Online localization combines averaged RSSI with the augmented map. Both solvers alternate ray casting, virtual-LOS compensation, and least-squares fitting; Algorithm~\ref{alg:ap_localization} and Section~\ref{sec:online_position} specify the updates.}
    \label{fig:pipeline}
\end{figure*}

We combine a deterministic propagation model with the wall geometry of \textit{OpenStreetMap Area Graph (osmAG)}~\cite{feng2023osmag}. Known anchor APs calibrate the signal model; unknown APs are localized offline and stored in the map for online robot position estimation. Our contributions are:
\begin{itemize}
\item \textbf{Structure-aware AP mapping:} joint use of signal measurements and osmAG geometry reduces the need to survey every AP manually.
\item \textbf{Generalization and efficiency:} mean error of 3.83~m in unfingerprinted areas versus 11.00--20.21~m for four WiFi baselines, with 251:1 compression of WiFi-related map data.
\item \textbf{Large-scale evaluation:} validation across 11,025~m$^2$ with narrow corridors, dense walls, and open atriums.
\end{itemize}

\section{RELATED WORK}
\subsection{Visual and LiDAR Global Localization}
Visual methods include Bag-of-Words matching and NetVLAD descriptors~\cite{galvez2012bags,arandjelovic2016netvlad}. Learned absolute pose regression also faces generalization limitations~\cite{sattler2019understanding}. LiDAR methods exploit 3D geometry yet face ambiguous corridors and expensive global registration~\cite{kim2018scan,aoki20243d}. WiFi provides a complementary global prior without requiring distinctive visual or geometric observations.

This complementarity is relevant to recovery after tracking failure: visually similar rooms and repeated corridor geometry can make place recognition ambiguous even when local tracking is accurate. We therefore target a coarse global initialization that can subsequently be refined by a higher-precision localizer, rather than replacing visual or LiDAR localization.

\subsection{WiFi-based Localization}
\label{sec:relatedwork_wifi_localization}
Fingerprinting matches RSSI against surveyed databases, commonly using $k$-nearest neighbors (KNN) or probabilistic models~\cite{bahl2000radar,youssef2005horus}. IRPLS-ABCL improves the radio map through interval lognormal-shadowing analysis and refines positions with adaptive Bayesian comprehensive-learning particle swarm optimization~\cite{10416396}. These approaches remain dependent on reference-data coverage and do not explicitly represent wall geometry.

Gaussian processes (GPs) model RSSI as a spatial function~\cite{ferris_gaussian_2006}. GP-PL fits per-AP log-distance priors and models their residuals with a variance-bounded GP~\cite{Miyagusuku2016}; GPoE combines per-AP posteriors using information-based weights~\cite{miyagusuku2018data}. Such models improve sparse-data interpolation but do not explicitly account for discrete wall attenuation. Our method instead uses osmAG ray casting to incorporate physical obstructions into AP mapping and robot localization.

The distinction is how each model handles environmental structure. Fingerprint enhancement improves the sampled radio map, and GP-based priors support interpolation through residual modeling. Explicit wall intersections instead condition the signal prediction on the path between a candidate position and an AP, including paths through areas without reference scans.

\section{METHOD}
The framework has offline calibration and online localization phases (Fig.~\ref{fig:pipeline}). Offline inputs are synchronized RSSI and LIO-SAM robot positions~\cite{shan2020lio}, anchor AP coordinates, and osmAG geometry. The learned propagation parameters and estimated AP locations augment the map; online localization combines this map with live RSSI.

\subsection{osmAG Map Representation}
\label{subsec:osmag-map}
osmAG stores geometry and topology in XML and can be constructed from CAD, grid, or point-cloud maps~\cite{feng2023osmag}. Coordinate nodes and polygonal ways represent room boundaries, walls, corridors, and their connections. Their explicit geometry supports ray-wall intersection tests.

\begin{figure}[t]
    \centering
    \includegraphics[width=0.98\linewidth]{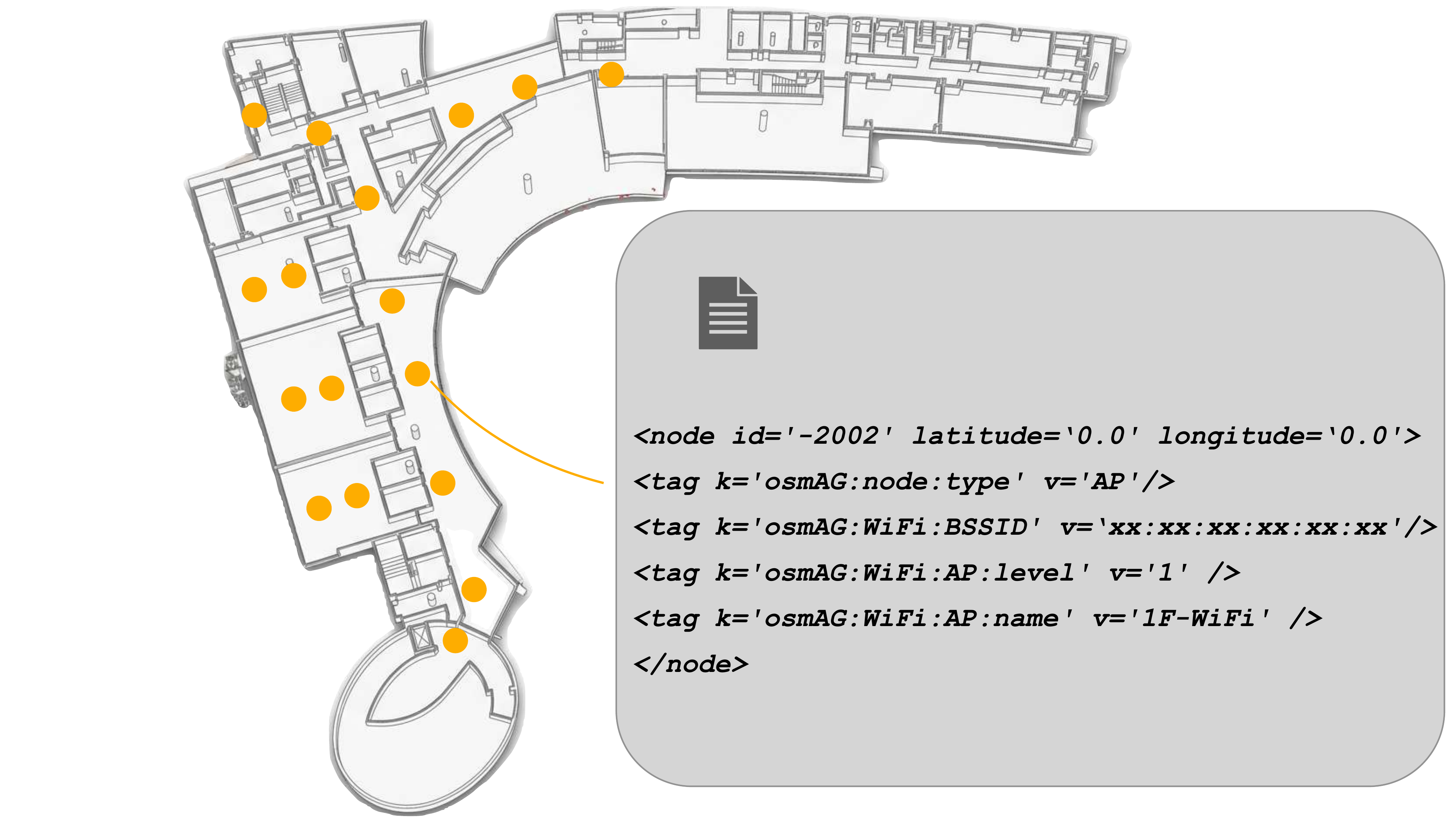}
    \caption{Augmented osmAG map: grey structures represent walls and rooms; orange nodes represent estimated APs. Semantic tags associate each AP's coordinates with its BSSID and floor level.}
    \label{fig:osmag_structure}
\end{figure}

We extend osmAG through semantic tags. Offline calibration nodes store RSSI and synchronized robot positions; mapped AP nodes store optimized 3D coordinates and attributes such as basic service set identifier (BSSID) and floor level (Fig.~\ref{fig:osmag_structure}). This unifies calibration data, AP priors, and geometry in one file. The signal model uses wall geometry, while osmAG retains topological and semantic information for navigation.

For calibration, the recorded robot position fixes one endpoint of each signal ray. For online localization, that endpoint is the current robot estimate and must be updated during optimization. Storing AP attributes beside the physical map lets both phases use the same geometric representation instead of maintaining a separate dense radio-map grid.

\subsection{Offline Phase: AP Localization}
Offline calibration estimates unknown AP positions and stores them in osmAG for subsequent robot localization.

\subsubsection{Structure-Aware Signal Propagation Model}
Known anchor APs calibrate the propagation model using only same-floor AP--observation pairs (Section~\ref{sec:param_learning}). We model RSSI as\vspace{-4mm}
\begin{equation}
    \text{RSSI} = \text{RSSI}_0 - 10n \log_{10}(d) - \sum_{k=1}^{N} W_k + X_{\sigma}
\end{equation}
where $\text{RSSI}_0$ is the reference strength at 1~m, $n$ is the path-loss exponent, $d$ is the 3D AP--robot distance, and $N$ counts intersected structural elements. Each contributes attenuation $W_k$; $X_\sigma$ is zero-mean Gaussian shadowing. To limit model complexity, we use one average wall-loss parameter $\overline{W}$, so $\sum_k W_k=N\overline{W}$.

The resulting model separates distance-dependent propagation from obstruction loss. The common attenuation factor is an effective average across wall materials, rather than a material-specific estimate. This simplification keeps the calibration compact while retaining an explicit dependence on the number of obstacles along each candidate signal path.

\subsubsection{Two-Stage Parameter Optimization}
\label{sec:two_stage_opti}
osmAG intersection tests separate line-of-sight (LOS) paths, with no intersected walls, from non-line-of-sight (NLOS) paths. We fit the parameters sequentially:

LOS observations first determine the distance-dependent baseline without a wall-loss term. Keeping that baseline fixed, the NLOS fit attributes the remaining systematic attenuation to intersected walls. This prevents the wall term from being omitted during calibration while avoiding simultaneous fitting of all propagation parameters from obstructed measurements alone.

\begin{enumerate}
    \item \textbf{LOS parameters.} Fit the reference strength and path-loss exponent on $\mathcal{S}_{\text{LOS}}$:
    \begin{equation}
    \label{eq2}
        (\text{RSSI}_0^*, n^*) = \arg \min_{\text{RSSI}_0, n} \sum_{(i,j) \in \mathcal{S}_{\text{LOS}}} e_{ij}^2
    \end{equation}
    \begin{equation}
        e_{ij} = \text{RSSI}_{ij} - \text{RSSI}_0 + 10n \log_{10}(d_{ij})
    \end{equation}
    where $\text{RSSI}_{ij}$ and $d_{ij}$ denote the measured RSSI and 3D distance for anchor AP $i$ and robot position $j$.
    \item \textbf{Wall attenuation.} Fix $(\text{RSSI}_0^*,n^*)$ and fit the wall loss on $\mathcal{S}_{\text{NLOS}}$:\vspace{-2mm}
    \begin{equation}
    \label{eq4}
        \overline{W} = \arg \min_{W} \sum_{(i,j) \in \mathcal{S}_{\text{NLOS}}} e_{ij}^2
    \end{equation}
    \begin{equation}
        e_{ij} = \text{RSSI}_{ij} - \text{RSSI}_0^* + 10n^* \log_{10}(d_{ij}) + N_{ij}W
    \end{equation}
    where $N_{ij}$ is the wall count for the AP--robot path.
\end{enumerate}

\subsubsection{Iterative AP Position Estimation}
With $\Theta=(\text{RSSI}_0^*,n^*,\overline{W})$ fixed, each unknown AP is assigned to its most frequently observed floor; only observations on that floor are retained. Algorithm~\ref{alg:ap_localization} initializes the AP position assuming LOS, then alternates geometry-based compensation and position fitting.

\begin{figure}[ht]
    \centering
    \subfloat[Initial position estimate\label{fig:iter_init}]{\includegraphics[width=0.48\linewidth]{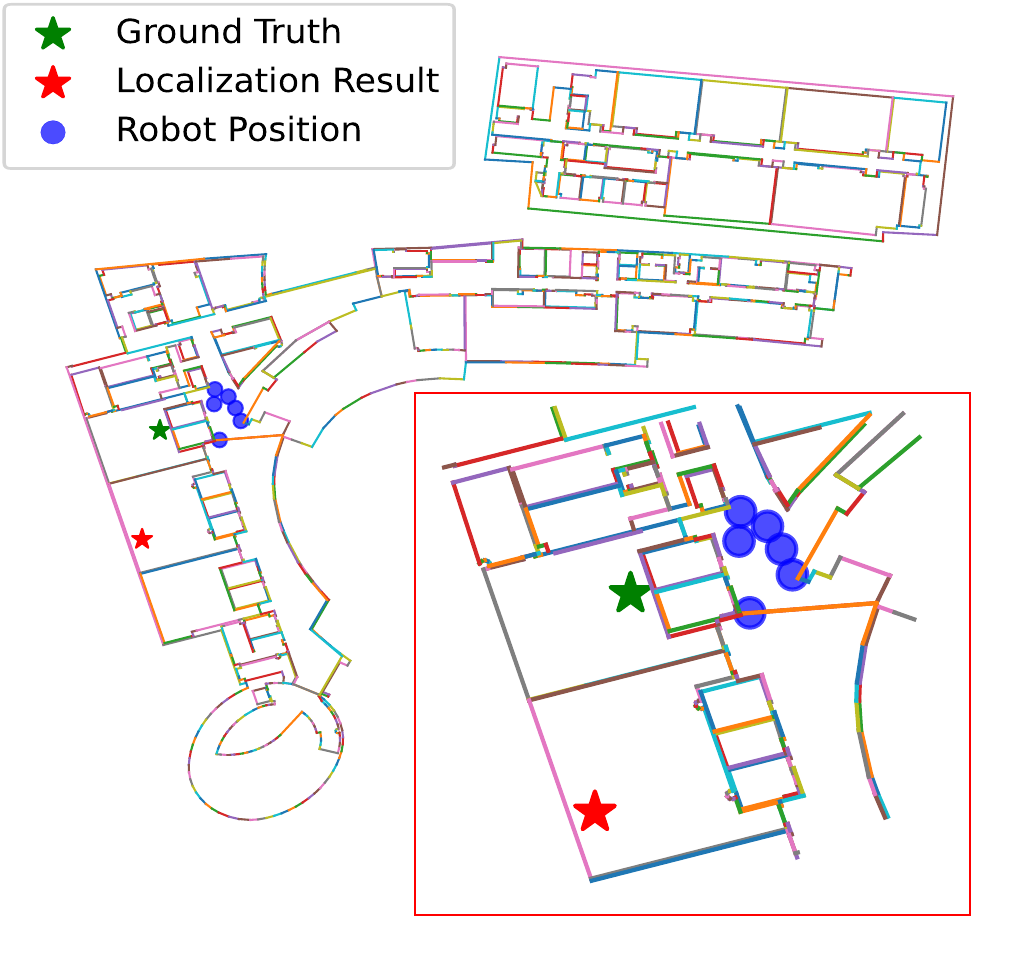}}
    \hfill
    \subfloat[Final position estimate\label{fig:iter_final}]{\includegraphics[width=0.48\linewidth]{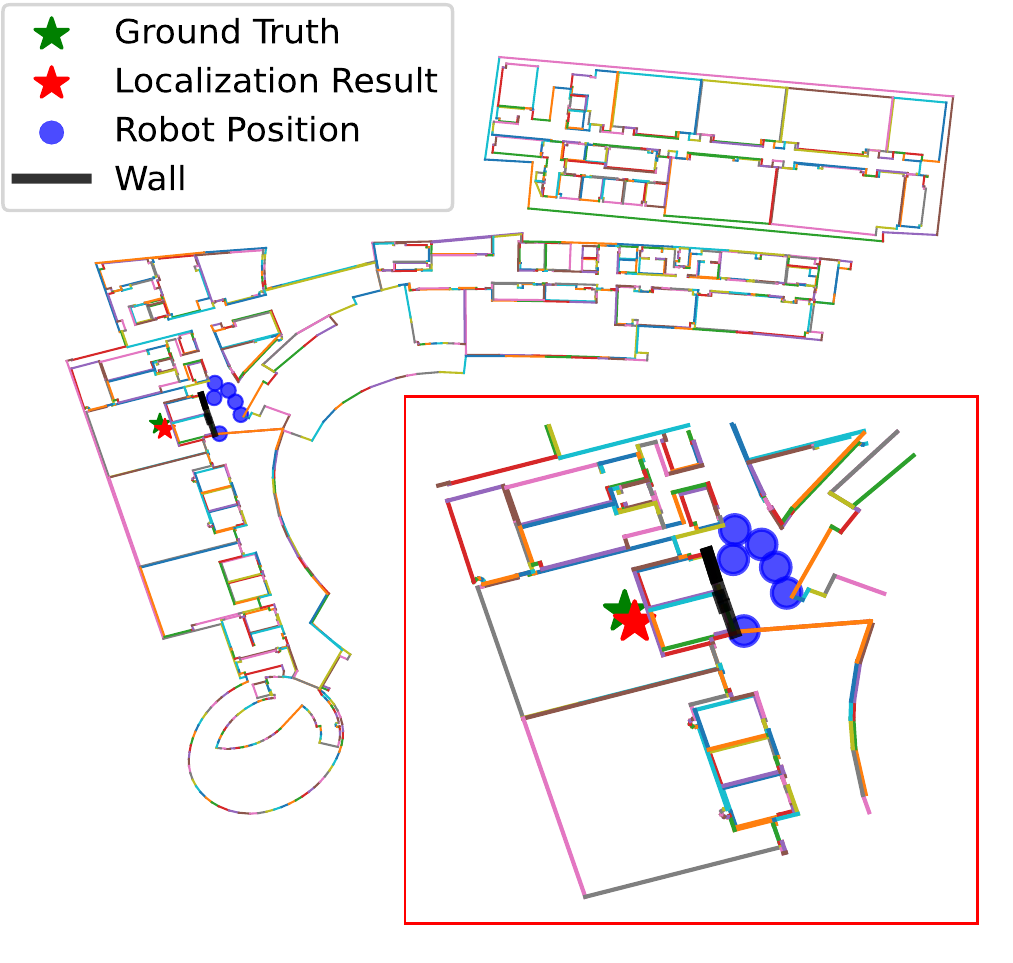}}
     \caption{Single-AP refinement. (a) LOS-only initialization. (b) Estimate after ten wall-aware iterations. Blue points are robot positions; the green star is AP ground truth. The red box marks the final-estimate zoom.}
    \label{fig:Iterative}
\end{figure}

\begin{algorithm}[t]
\caption{Structure-Aware Iterative AP Localization}
\label{alg:ap_localization}
\begin{algorithmic}[1]
\Require Same-floor measurements $\mathcal D=\{(\mathbf p_i,\text{RSSI}_i)\}_{i=1}^{M}$; wall polygons $\mathcal M$; parameters $\Theta$; $T_{\max}$ and $\epsilon$
\Ensure Estimated 3D AP position $\mathbf p_{\mathrm{AP}}$
\State \textbf{LOS-only initialization}
\State $\hat d_i^{(0)}\gets10^{(\text{RSSI}_0^*-\text{RSSI}_i)/(10n^*)}$ for all $i$
\State $\mathbf p_{\mathrm{AP}}^{(0)}\gets\arg\min_{\mathbf p}\sum_i(\|\mathbf p-\mathbf p_i\|-\hat d_i^{(0)})^2$
\State \textbf{Structure-aware refinement}
\For{$t\gets1$ to $T_{\max}$}
  \For{each measurement $i$}
    \State $N_i\gets\text{CountWalls}(\text{LineString}(\mathbf p_{\mathrm{AP}}^{(t-1)},\mathbf p_i),\mathcal M)$
    \State $\text{RSSI}'_i\gets\text{RSSI}_i+N_i\overline W$ \Comment{unchanged if LOS}
    \State $\hat d_i^{(t)}\gets10^{(\text{RSSI}_0^*-\text{RSSI}'_i)/(10n^*)}$
  \EndFor
  \State $\mathbf p_{\mathrm{AP}}^{(t)}\gets\arg\min_{\mathbf p}\sum_i(\|\mathbf p-\mathbf p_i\|-\hat d_i^{(t)})^2$
  \If{$\|\mathbf p_{\mathrm{AP}}^{(t)}-\mathbf p_{\mathrm{AP}}^{(t-1)}\|<\epsilon$}
    \State \textbf{break}
  \EndIf
\EndFor
\State \Return $\mathbf p_{\mathrm{AP}}^{(t)}$

\end{algorithmic}
\end{algorithm}
Ignoring NLOS attenuation makes RSSI-derived distances too large, biasing the initial estimate (Fig.~\ref{fig:Iterative}a). At each refinement, ray casting updates wall counts from the current AP estimate, and adding $N_i\overline W$ to RSSI produces a \textit{virtual LOS} measurement. Least-squares fitting then updates the AP position and hence the next wall counts. Refinement stops when the position change is below $\epsilon$ or $T_{\max}$ is reached. Fig.~\ref{fig:Iterative}b shows the estimate after ten iterations.

The magnitude of this bias follows directly from the propagation model. Ignoring shadowing, LOS-only inversion gives $\hat d_{\mathrm{LOS}}=d\,10^{N\overline W/(10n)}$. With the learned parameters, one intersected wall would therefore inflate the distance by a factor of approximately 2.7 before compensation. This is a model prediction, not an additional measurement. It explains how wall loss can substantially distort trilateration even when the RSSI observations themselves are stable.

The wall counts cannot be fixed before an unknown AP has been localized: moving the candidate AP changes which polygons the signal rays intersect. Conversely, those intersections change the RSSI-to-distance conversion used by the position solver. Alternation couples these two quantities while keeping the globally learned propagation parameters fixed. When a ray intersects no walls, the compensation is zero and the original LOS distance model is recovered. The stopping test detects small position updates; it does not establish a global optimum.

\subsection{Online Phase: Robot Position Estimation}
\label{sec:online_position}
The robot averages RSSI over a short window (e.g., 3--5~s). Its floor is inferred by an unweighted vote over distinct detected BSSIDs with mapped floor labels; only observations on the winning floor are retained. Ray casting uses that floor's 2D plan. This handles cross-floor reception through floor selection and filtering, without an explicit floor-slab attenuation model.

LOS-only trilateration provides an initial estimate of the robot's position. Refinement repeats the AP-localization compensation step using rays from the current robot estimate to mapped APs, then solves

\begin{equation}
        \mathbf{p}_{\text{r}}^{(t)} = \arg \min_{\mathbf{p}} \sum_{k=1}^{K} \left( \| \mathbf{p} - \mathbf{p}_k^{\text{AP}} \| - \hat{d}_k^{(t)} \right)^2
    \end{equation}
where $\mathbf p_k^{\mathrm{AP}}$ is a selected AP's map position and $\hat d_k^{(t)}$ is its compensated distance. The experiments use ten position-optimization steps: one initialization and nine refinements.

The selected AP coordinates and propagation parameters are map priors, while only the robot position is updated online. Compensation is recomputed from the current estimate at each refinement, so the robot solver also adapts its wall counts as the candidate position moves. The fixed experimental budget makes computation comparable across test positions.

Selecting the strongest APs does not explicitly optimize geometric diversity. When selected APs are nearly collinear, several candidate robot positions may produce similar range residuals. Wall-aware compensation corrects the propagation model, but cannot by itself remove every ambiguity caused by the AP layout.

\section{EXPERIMENTS}
\providecommand{\blind}[1]{{}}

\subsection{Experimental Setup}
\begin{figure}[ht]
    \centering
    \subfloat[Floor 1\label{fig:env_floor1}]{\includegraphics[width=0.48\linewidth]{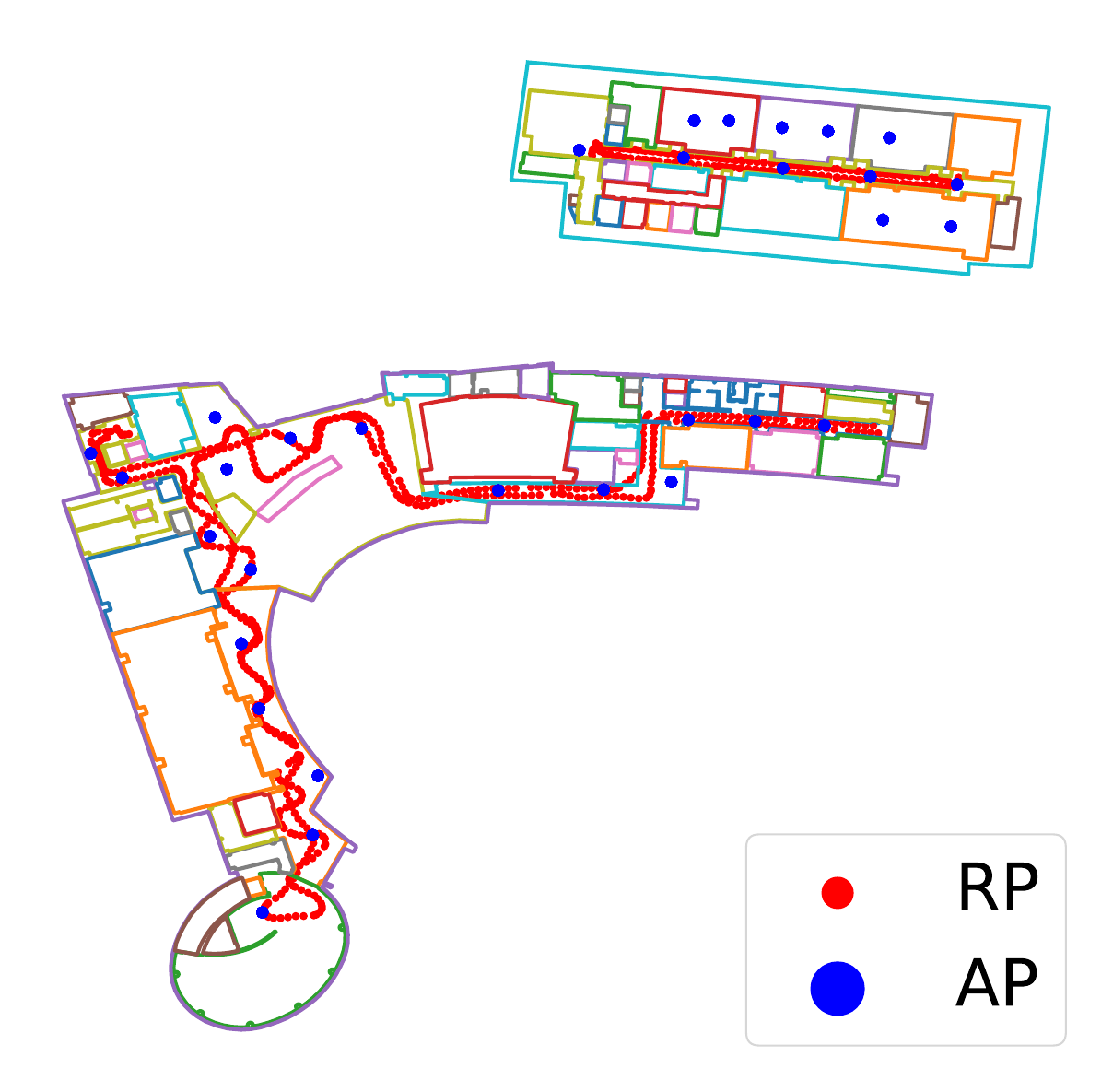}}
    \hfill
    \subfloat[Floor 2\label{fig:env_floor2}]{\includegraphics[width=0.48\linewidth]{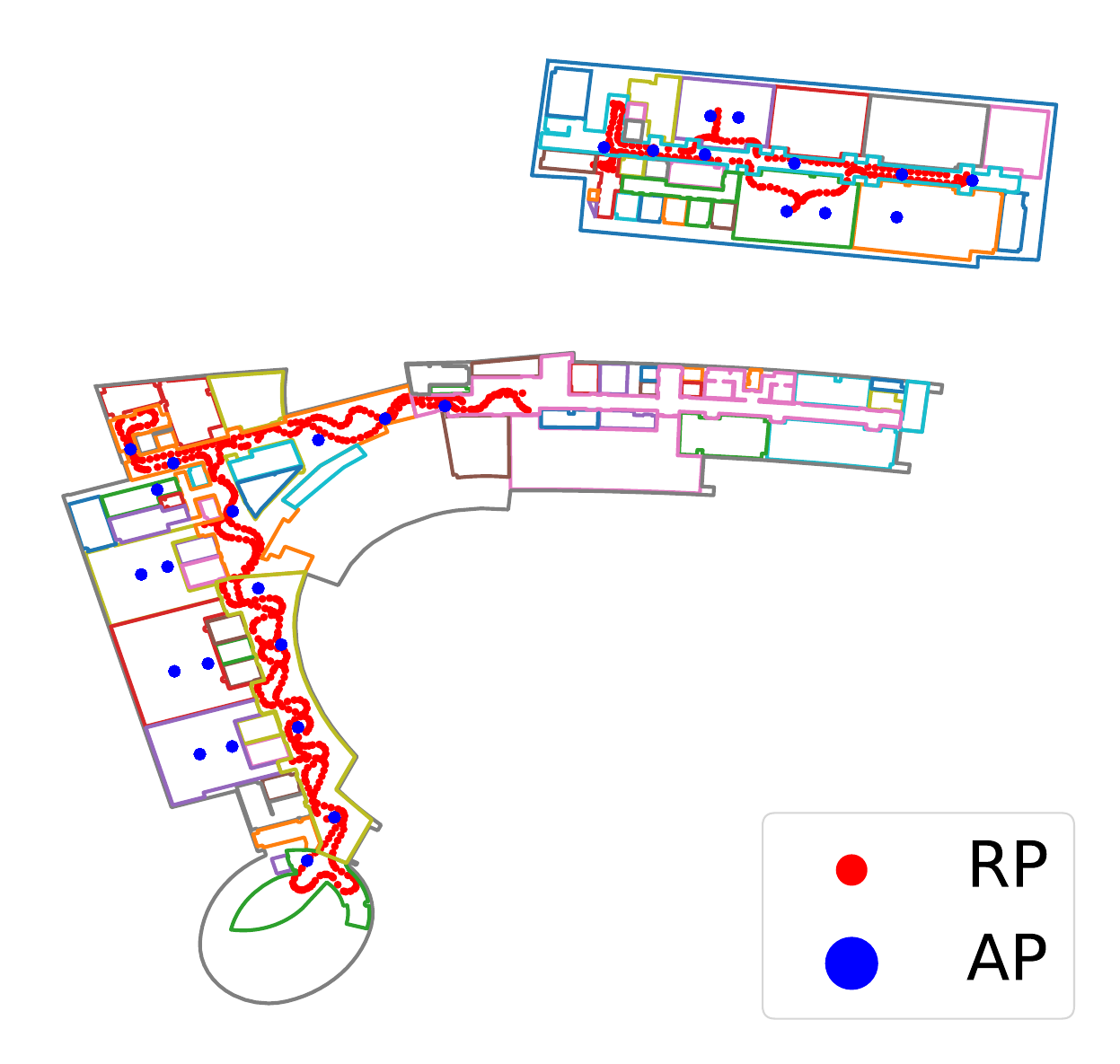}}
   
    \caption{Experimental osmAG layouts across Buildings A and B: (a) Floor~1; (b) Floor~2. Red RP markers indicate robot positions, and blue AP markers indicate surveyed AP locations.}
    \label{fig:env}
\end{figure}

The 11,025~m$^2$ test site spans two floors across two campus buildings (Fig.~\ref{fig:env}). Building A has narrow corridors and dense partitions; Building B has open spaces and atriums. The site includes classrooms, labs, and offices with concrete, drywall, and glass walls. An Agilex Hunter SE carried an EDUP EP-AC1681 WiFi receiver and a Hesai 64-channel LiDAR. LIO-SAM provided robot ground truth~\cite{shan2020lio}; all 49 AP locations were manually surveyed.

\begin{figure}[t]
    \centering
    \includegraphics[width=0.95\linewidth]{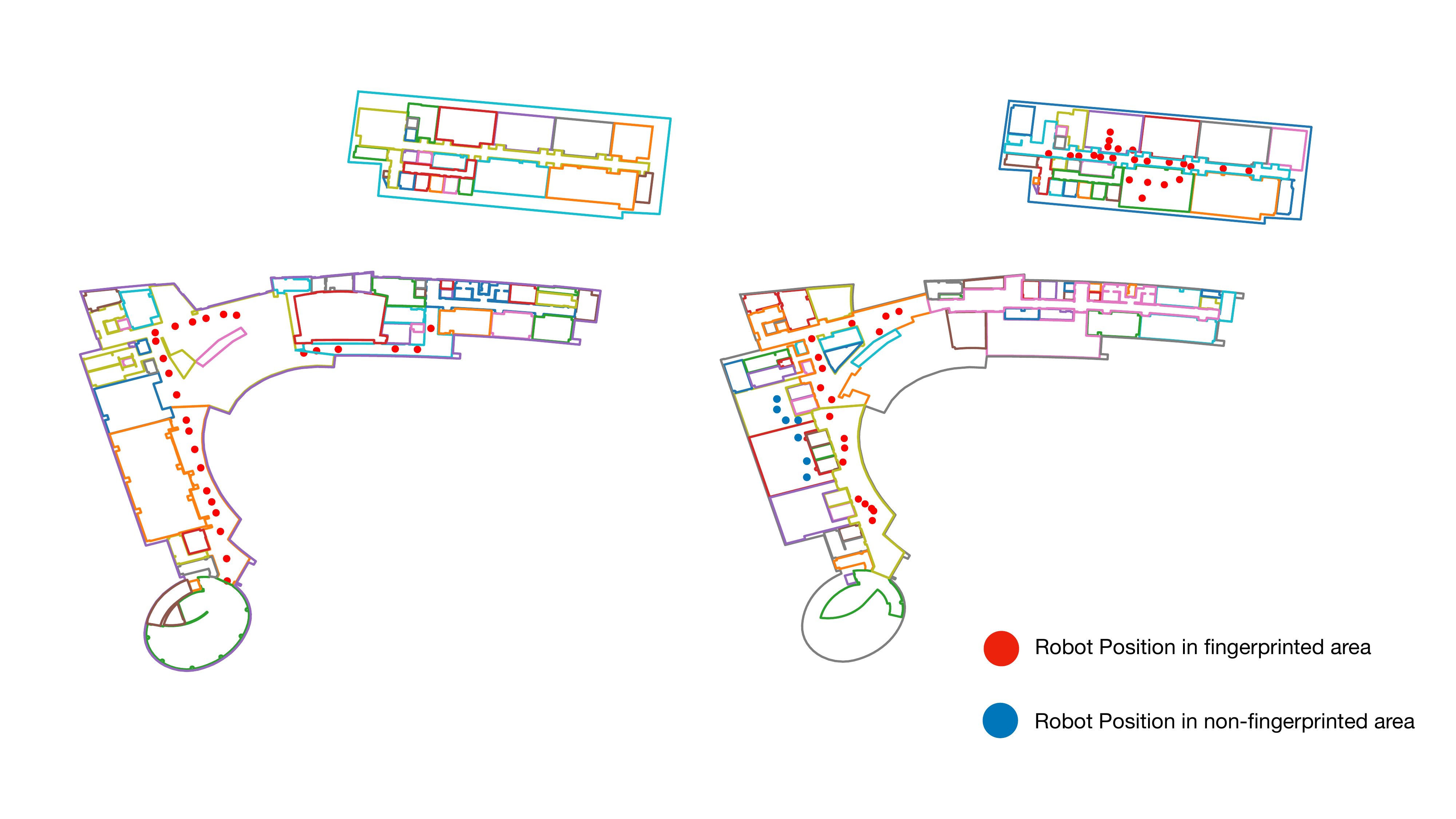}
    \caption{Robot test positions across two floors: 66 fingerprinted positions (red) and 7 unfingerprinted positions (blue), including cross-building corridors and atriums.}
    \label{fig:robot_test_positions}
\end{figure}

Eight APs serve as calibration anchors, and the remaining 49 are used to evaluate offline AP localization. Robot localization uses 66 positions in fingerprinted areas and 7 in unfingerprinted areas (out-of-distribution, OOD); Fig.~\ref{fig:robot_test_positions} shows their distribution.

The two buildings expose the model to contrasting propagation conditions: frequent wall crossings in partitioned corridors and longer unobstructed paths in open spaces. The fingerprinted and OOD robot sets assess localization within surveyed coverage and beyond that coverage, respectively; the OOD positions are separate spatial tests, not repeated samples of the fingerprinted set. Metrics are mean error, standard deviation (Std.), root-mean-square error (RMSE), and the 95th percentile (P95).

\begin{table*}[htbp]
\centering
\caption{Quantitative Comparison of Robot Localization Performance Across Different Scenarios.}
\label{tab:robot_comparison_sota}
\small
\renewcommand{\arraystretch}{1.1}
\setlength{\tabcolsep}{4pt}
\begin{tabular}{@{}lcccccccc@{}}
\toprule
\multirow{3}{*}{\textbf{Method}} & \multicolumn{4}{c}{\textbf{Fingerprinted Areas} (In-Distribution)} & \multicolumn{4}{c}{\textbf{Unfingerprinted Areas} (Out-of-Distribution, N=7)} \\
\cmidrule(lr){2-5} \cmidrule(l){6-9}
 & \textbf{Mean} & \textbf{Std.} & \textbf{RMSE} & \textbf{P95} & \textbf{Mean} & \textbf{Std.} & \textbf{RMSE} & \textbf{P95} \\
 & (m) & (m) & (m) & (m) & (m) & (m) & (m) & (m) \\
\midrule
KNN Fingerprinting & 3.42 & 2.50 & 4.24 & 7.85 & 20.21 & 9.48 & 22.32 & 36.18 \\
GP-PL~\cite{Miyagusuku2016} & 3.18 & 2.81 & 4.24 & 6.50 & 13.70 & 11.50 & 17.89 & 22.90 \\
GPoE~\cite{miyagusuku2018data} & 3.12 & 1.77 & 3.59 & 6.63 & 14.71 & 2.67 & 14.95 & 18.72 \\
IRPLS-ABCL~\cite{10416396} & \textbf{3.05} & 1.88 & 3.59 & 6.63 & 11.00 & 2.41 & 11.26 & 13.65 \\
\textbf{Ours (osmAG-Structure)} & 3.12 & \textbf{1.35} & \textbf{3.40} & \textbf{4.65} & \textbf{3.83} & \textbf{0.78} & \textbf{3.91} & \textbf{4.72} \\
\bottomrule
\end{tabular}
\end{table*}

\subsection{Parameter Learning}
\label{sec:param_learning}
The eight anchors on the first floor of Building A calibrate Eqs.~\ref{eq2} and \ref{eq4}, yielding $\text{RSSI}_0^*=-28.79$~dBm, $n^*=2.5$, and $\overline W=10.77$~dB. These parameters remain fixed for AP and robot localization. Calibration anchors are excluded from both localization evaluations.

The training area contains the dense partitions for which structural compensation is intended. Once learned, the same parameter values are applied across the evaluation site without refitting them for each unknown AP or robot query. Surveyed coordinates for non-anchor APs serve as evaluation ground truth, not localization inputs.

\subsection{Access Point Localization}
\label{sec:ap_localization}
\begin{figure}[t]
    \centering
    \includegraphics[width=0.95\linewidth]{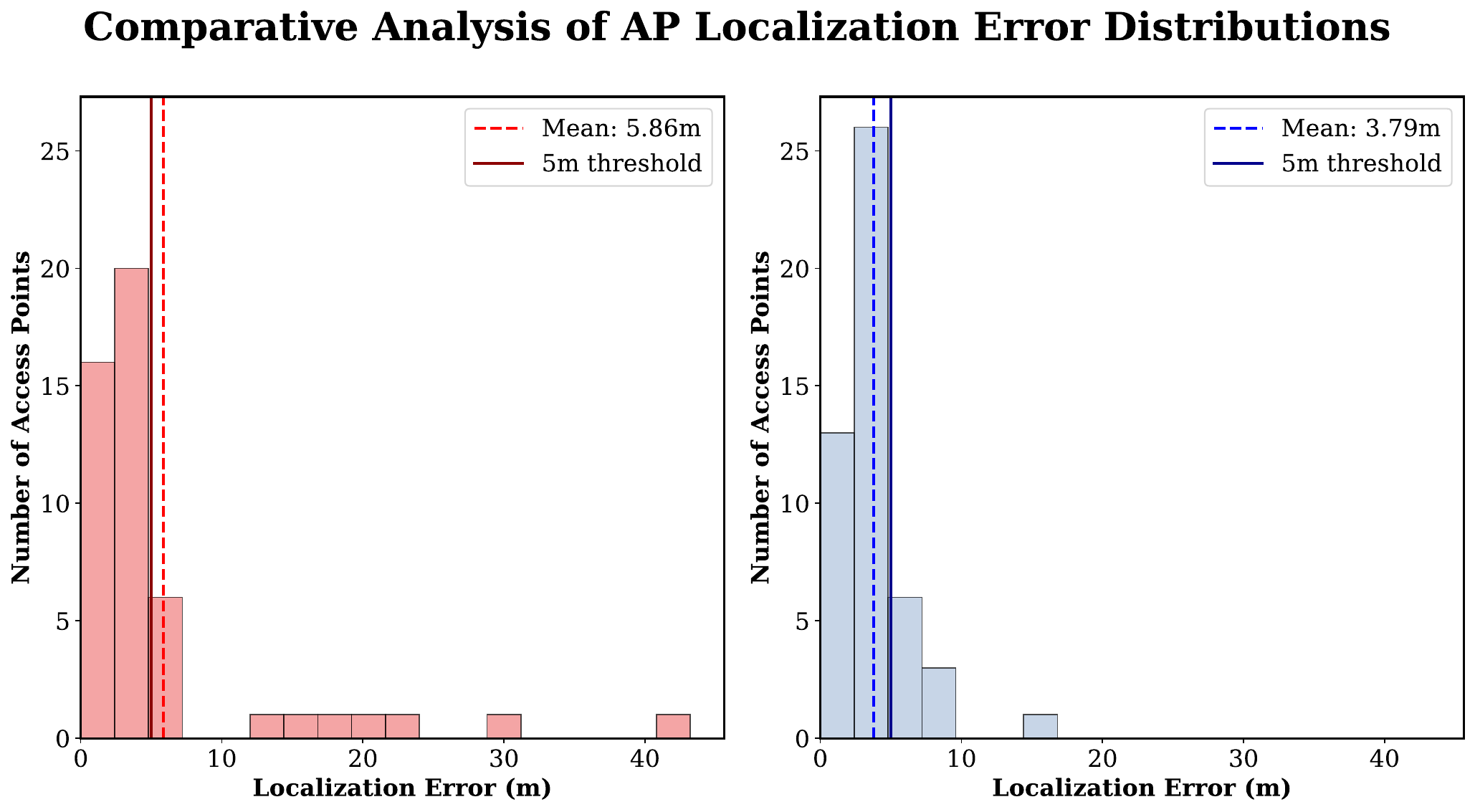}
    \caption{Per-AP errors for 49 non-anchor APs: geometry-unaware trilateration versus structure-aware refinement.\vspace{-4mm}}
    \label{fig:ap_comparison}
\end{figure}

Across the 49 non-anchor APs, geometry-unaware trilateration yields a mean error of 5.86~m and standard deviation of 7.82~m. Structure-aware refinement reduces these to \textbf{3.79~m} and 2.52~m: a \textbf{35.3\%} mean-error reduction. Fig.~\ref{fig:ap_comparison} shows the tighter error distribution after accounting for NLOS attenuation.

The reduction in standard deviation is also relevant: the geometry-unaware baseline has a much wider spread of AP errors than the wall-aware solution. Jointly inspecting the mean and the per-AP distribution avoids characterizing calibration quality by a single typical case. The single-AP example in Fig.~\ref{fig:Iterative} illustrates the update mechanism; Fig.~\ref{fig:ap_comparison} evaluates it across the full non-anchor set.

\subsection{Robot Localization}
The evaluator uses the estimated AP map, excludes anchors, and selects the three strongest distinct mapped AP locations without substituting surveyed coordinates. It uses recorded floor annotations and does not run online floor voting or strict same-floor filtering. Thus, results assess position estimation with supplied floor information, not the complete online floor-inference pipeline. Selected links include cross-floor APs at 7 of 66 fingerprinted positions; all 7 OOD positions use same-floor APs. AP and robot errors concern different targets and test sets, so their means do not define a direct error-propagation ratio; robot errors also depend on AP geometry, RSSI, and map constraints.

AP coordinate errors enter query-specific range residuals, whose combined effect on the position estimate depends on the selected AP geometry and the other RSSI observations.

We compare KNN fingerprinting, GP-PL, GPoE, and IRPLS-ABCL (Section~\ref{sec:relatedwork_wifi_localization}). Table~\ref{tab:robot_comparison_sota} reports all metrics.

KNN represents dense database matching; GP-PL uses path-loss-informed spatial interpolation; GPoE fuses multiple AP-specific predictions; IRPLS-ABCL combines radio-map enhancement with particle swarm optimization (PSO) refinement. These baselines cover different uses of surveyed RSSI, allowing the fingerprinted and OOD results to distinguish interpolation performance from generalization beyond the reference data.

In fingerprinted areas, mean errors are similar (3.05--3.42~m). Our mean is 3.12~m, with the lowest standard deviation (1.35~m) and 95th-percentile error (4.65~m), indicating a tighter error distribution.

The improvement in this setting concerns dispersion and tail error, not a uniformly lower mean: IRPLS-ABCL has a slightly smaller mean, and GPoE has the same reported mean as our method. The full statistics therefore support a more consistent estimate within surveyed coverage rather than superiority on every metric.

In unfingerprinted areas, our mean error is \textbf{3.83~m} (standard deviation 0.78~m; 95th percentile 4.72~m), versus 20.21~m for KNN, 13.70~m for GP-PL, 14.71~m for GPoE, and 11.00~m for IRPLS-ABCL. Explicit wall modeling supports localization beyond fingerprint coverage. However, the OOD set contains only seven positions, including cross-building corridors and atriums; broader validation is needed.

The OOD comparison is the more direct test of extrapolation because no reference fingerprint is available at those positions. The low errors on this test set are consistent with using mapped wall geometry to constrain signal predictions outside the survey footprint. They do not establish performance for every building layout or validate autonomous floor recognition, which the recorded-floor evaluation does not exercise.

\subsection{Computational Cost}
\label{sec:computational_cost}
With single-threaded numerical libraries, the solver averaged 278.2~ms per query (P95: 343.9~ms; Table~\ref{tab:runtime}). After warm-up, all 73 test positions were evaluated five times, using three APs and one initialization plus nine refinement solves per query. Timing includes wall-intersection checks, signal compensation, and position fitting, but excludes data acquisition, preprocessing, and online floor inference. The reported runtime measures solver computation rather than end-to-end latency or continuous localization throughput.

\begin{table}[t]
\centering
\caption{Per-query solver computation time over five repeated passes.}
\label{tab:runtime}
\small
\begin{tabular}{@{}lrrr@{}}
\toprule
\textbf{Test set} & \textbf{Mean (ms)} & \textbf{P95 (ms)} & \textbf{Max. (ms)} \\
\midrule
Fingerprinted Areas & 276.9 & 341.7 & 394.6 \\
Unfingerprinted Areas & 289.9 & 370.6 & 381.5 \\
All positions & 278.2 & 343.9 & 394.6 \\
\bottomrule
\end{tabular}
\end{table}

\subsection{Storage Efficiency}
\label{sec:storage_efficiency}
Table~\ref{tab:storage} compares WiFi-related storage with the base osmAG geometry unchanged. The augmented representation achieves approximately \textbf{251:1 compression} relative to the raw fingerprint database, reducing the storage burden of large-scale deployment.

The comparison concerns WiFi-related records: 1,491 raw scans occupy 11,068~KB, whereas the 49-AP representation occupies 44~KB. It is not a claim that the complete geometric map is compressed by the same factor. Retaining a compact AP layer avoids storing a dense fingerprint database for localization while preserving the underlying semantic map.

\begin{table}[h]
\centering
\caption{Storage Efficiency Comparison of WiFi Map Representations.}
\label{tab:storage}
\footnotesize
\renewcommand{\arraystretch}{1.1}
\setlength{\tabcolsep}{4pt}
\begin{tabular}{@{}lccc@{}}
\toprule
\textbf{Representation} & \textbf{Points} & \textbf{Size (KB)} & \textbf{Per-Unit} \\
\midrule
Fingerprint DB & 1,491 scans & 11,068 & 7.42 KB/scan \\
osmAG + AP (Ours) & 49 APs & 44 & $<$1 KB/AP \\
\bottomrule
\end{tabular}
\end{table}

\section{CONCLUSIONS}
Combining osmAG wall geometry with signal propagation modeling improves AP calibration and robot localization beyond fingerprint coverage while retaining a compact map. The method achieves competitive accuracy in surveyed areas and lower errors than four WiFi baselines at the seven unfingerprinted test positions. Future work will evaluate the method on larger OOD sets, test the complete online floor-inference pipeline, and integrate this global initializer with LiDAR-based refinement~\cite{xie2023agloc} and osmAG navigation~\cite{11346008,11246069}.

The intended role is a scalable global initializer over large indoor areas. A complete deployment must additionally evaluate floor inference, RSSI acquisition, and downstream pose refinement together; the present solver timing and recorded-floor accuracy experiments assess only parts of that complete system.

\bibliographystyle{IEEEtran}

\bibliography{references}

\end{document}